\documentclass[letterpaper, 10 pt, conference]{ieeeconf}  

\IEEEoverridecommandlockouts                              
\usepackage{graphics}
\usepackage{epsfig}
\usepackage{amsmath}
\usepackage{amssymb}

\usepackage[ruled, linesnumbered]{algorithm2e}
\usepackage{subfigure}
\usepackage{stfloats}
\usepackage{adjustbox}

\usepackage{cite}
\usepackage{soul}
\usepackage{changes}
\usepackage{cancel}

\newcommand{\argmin}{\mathop{\mathrm{argmin}}}

\usepackage{hyperref}
\hypersetup{
    colorlinks=true,
    linkcolor=blue,      
    urlcolor=blue
}

\usepackage{xcolor}
\title{\LARGE \bf
Computationally and Sample Efficient Safe Reinforcement Learning  \\
Using Adaptive Conformal Prediction
}

\author{Hao Zhou, Yanze Zhang, and Wenhao Luo%
\thanks{$^*$This work was supported in part by the U.S. National Science Foundation under Grant CNS-2312465.}
\thanks{Authors are with the Department of Computer Science, University of Illinois Chicago, Chicago, IL 60607, USA.
Email: {\tt\small \{hzhou134, yzhan361, wenhao\}@uic.edu}%
}
}

\begin{document}

\maketitle
\thispagestyle{empty}
\pagestyle{empty}

\begin{abstract}
Safety is a critical concern in learning-enabled autonomous systems especially when deploying these systems in real-world scenarios. An important challenge is accurately quantifying the uncertainty of unknown models to generate provably safe control policies that facilitate the gathering of informative data, thereby achieving both safe and optimal policies. Additionally, the selection of the data-driven model can significantly impact both the real-time implementation and the uncertainty quantification process. In this paper, we propose a provably sample efficient episodic safe learning framework that remains robust across various model choices with quantified uncertainty for online control tasks. Specifically, we first employ Quadrature Fourier Features (QFF) for kernel function approximation of Gaussian Processes (GPs) to enable efficient approximation of unknown dynamics. Then the Adaptive Conformal Prediction (ACP) is used to quantify the uncertainty from online observations and combined with the Control Barrier Functions (CBF) to characterize the uncertainty-aware safe control constraints under learned dynamics. Finally, an optimism-based exploration strategy is integrated with ACP-based CBFs for safe exploration and near-optimal safe nonlinear control. Theoretical proofs and simulation results are provided to demonstrate the effectiveness and efficiency of the proposed framework.
\end{abstract}

\section{Introduction}

Model-based Reinforcement Learning (MBRL) \cite{chowdhury2019online,chua2018deep,curi2020efficient,deisenroth2011pilco,kamthe2018data} has shown promising results when applied to variant nonlinear systems because of its robust generalization capabilities. However, the inherent exploratory nature of MBRL and the presence of partially or completely unknown dynamics pose significant challenges when deploying these methods in safety-critical environments. For example, indiscriminate exploration strategies, such as random exploration, can cause irreversible damage to mechanical systems like robots before any meaningful learning takes place in real-world scenarios. This highlights the need for developing a safety-critical learning framework to enable safe exploration, ensuring that data collection for learning the system’s dynamics does not compromise the safe operation of the system. 

To guarantee safety during the learning process, it is necessary to accurately model the system dynamics under conditions of uncertainty.
GPs \cite{cao2017gaussian,cheng2019end,koller2018learning,wang2018safe,greeff2020exploiting} are often employed for this purpose because they can model complex, nonlinear functions without being restricted to a specific parametric form, while also providing confidence intervals that capture model uncertainty \cite{curi2020efficient,koller2018learning}. Despite these capabilities, incorporating explicit safety constraints within the learning framework remains a critical challenge. Recent advancements have addressed this by integrating the uncertainty quantification derived from GPs into established control frameworks.
For example, work in \cite{koller2018learning} incorporates GPs-based uncertainty regions into a Model Predictive Control (MPC) framework to facilitate safe learning. Similarly, work in \cite{cheng2019end,wang2018safe} utilizes the uncertainty bounds from GPs to define safety constraints through Control Barrier Functions \cite{ames2019control} using the characteristics of CBF to enforce the safety set forward invariant. However, the assumption that the noise follows a Gaussian distribution or uniform distribution \cite{luo2020multi} can restrict their applicability in scenarios where the noise exhibits more complex distribution characteristics. 
Moreover, changes in the distribution of input features can lead to a covariance shift, complicating the accurate quantification of safety constraints. Even though Adaptive Conformal prediction (ACP) \cite{dixit2023adaptive, zaffran2022adaptive,zhou2024safety} shows potential in addressing the covariance shift issue, how to integrate this technique into safe learning framework to guide exploration remains an unresolved challenge.

The second challenge is achieving data efficiency, i.e., learning effective policies with a relatively small amount of data or interactions with the environment. The key is to strike a balance between exploration, which aims to gather informative data, and exploitation, which focuses on utilizing the collected data for system control. One common strategy to address this challenge is the integration of the Upper Confidence Bound (UCB) with GPs, known as GP-UCB \cite{srinivas2009gaussian}. The resulting exploration-exploitation trade-off improves the dynamics model learning and provides sample-efficiency guarantees for GPs-based algorithms \cite{srinivas2009gaussian}. Besides, the work in \cite{curi2020efficient} employs an optimistic exploration strategy using UCB to optimize both policies and models concurrently, achieving sublinear regret in certain scenarios. Thompson Sampling (TS) \cite{dumitrascu2018pg,russo2018tutorial} is another significant strategy for balancing exploration and exploitation. It operates by sampling a reward distribution for each action and selecting the action with the highest sampled reward, which allows for a more probabilistic approach to tackling uncertainty. Work in \cite{luo2022sample} apply TS in conjunction with linear regression within a robotic context, demonstrating computational and sample efficiency.
\begin{figure}[t]
	\centering
		\centering
		\includegraphics[scale=1, width=0.65\linewidth]{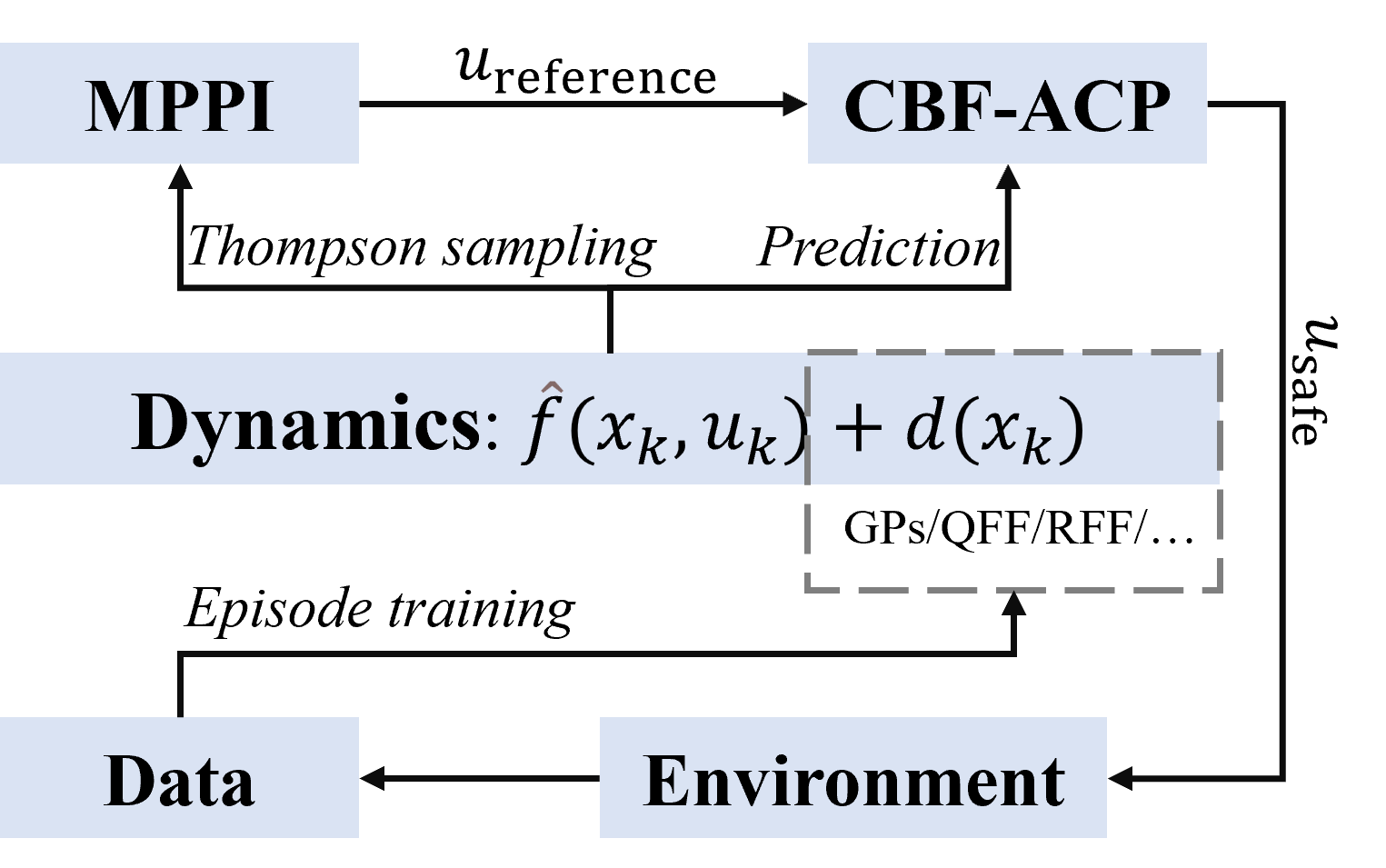}
	\caption{Computationally and sample efficient safe learning framework. The dynamics by Thompson sampling is embedded into the Model Predictive Path Integral (MPPI)\cite{williams2018information,williams2017information} to generate the reference policy. Then, ACP-based CBF is applied to guarantee safety. The framework is performed episodically.
    }
	\label{fig: framework}
\end{figure}

The combination of GP-UCB with safety constraints \cite{curi2020efficient,wang2018safe} shows a potential to achieve the sample efficient safe learning under conditions of uncertainty, particularly when assuming Gaussian-distributed noises. However, posterior sampling of the GP-UCB to obtain the maximum uncertain exploration state is time-consuming because the prediction complexity of GPs is $O(N^3)$ where $N$ is the number of data points. This poses a substantial bottleneck for real-time applications in online safe learning tasks. To address this computational challenge, Random Fourier Features (RFF) have been employed to approximate GP kernel functions, facilitating online sample-efficient learning and enabling real-time control of nonlinear systems. Nevertheless, a critical drawback of RFF is its inability to provide reliable confidence bounds as shown in \cite{mutny2018efficient}. Even though Quadrature Fourier Features \cite{mutny2018efficient} demonstrates superior error approximation capabilities, allowing for accurate approximations of both the posterior mean and variance, how to integrate QFF into a safety-critical learning scheme for online nonlinear control remains challenging. 

To realize the online safe learning task involved with the computational and sample efficiency, we propose to utilize the QFF for kernel function approximation of GPs to efficiently and precisely approximate unknown dynamics. 
Subsequently, we utilize Adaptive Conformal Prediction to accurately quantify the uncertainty of CBF within the framework of the learned dynamics, facilitating safe learning procedures. Moreover, we integrate an optimism-based exploration strategy with ACP-based CBF, enabling safe exploration that leverages the learned dynamics for achieving near-optimal safe control performance. Our contributions are as follows:
\begin{itemize}
\item We present a novel method based on Adaptive Conformal Prediction to quantify the uncertainty of the safety constraint during online safe learning, providing a robust and versatile tool for enhancing safety with learned system dynamics;
\item We introduce a computationally and sample efficient episodic online learning framework (Fig.\ref{fig: framework}) that leverages the approximated GPs with the optimism-based exploration strategy to simultaneously achieve safe learning and policy optimization for online nonlinear control tasks;
\item We provide theoretical analysis and experiment results on different platforms to demonstrate the performance of our proposed approach.
\end{itemize}

\section{Preliminary}
\subsection{Quadrature Fourier Features}
Quadrature Fourier Features \cite{mutny2018efficient} is able to handle complex kernel functions in a computationally efficient manner, making them particularly useful in machine learning algorithms such as GPs. Unlike RFF to approximate the GPs, QFF adopts the quadrature integration to achieve the precise approximation of the kernel. In this paper, we represent the QFF as $\phi(X)$ where $X=(x, u)$. $x\in \mathbb{R}^n$ is the robot state and $u \in \mathbb{R}^m$ is the control input. The specific form of $\phi(X)$  and its' approximation error relative to the GPs can be found in Definition 3 from \cite{mutny2018efficient} and Theorem 1 from \cite{mutny2018efficient}, respectively.

\subsection{Adaptive Conformal Prediction} \label{sec: ACP}
Conformal Prediction (CP) \cite{papadopoulos2002inductive} is a statistical method that formulates a certified region of prediction for complex prediction models to cover the ground truth value with guaranteed probability, and is independent of prediction model choice. 
Given a dataset $\mathcal{D}=\{(X_k, Y_k)\}$ (where $k=1,\dots, \overline{K}$, $X_k \in \mathbb{R}^n$, and $Y_k \in \mathbb{R}$) and any prediction model $f^c: X \mapsto Y$ trained from $\mathcal{D}$, e.g. linear model or deep learning model, the goal of CP is to obtain $\Bar{S} \in \mathbb{R}$ to construct a region $R(X^{*})=[f^c(X^{*})-\Bar{S}, f^c(X^{*})+\Bar{S}]$ so that $\mathrm{Pr}(Y^{*} \in R(X^{*})) \geqslant 1-\alpha^c$, where $\alpha^c \in (0, 1)$ is the failure probability and $X^{*}, Y^{*}$ are the testing model input and the true value respectively. We can then define the nonconformity score $S_{k} \in \mathbb{R}^{+}$ on calibration dataset $\mathcal{T}=\{(\Bar{X}_k, \Bar{Y}_k)\}$, which is computed by $S_{k} = |\Bar{Y}_k -  f^c(\Bar{X}_k)|$. The large nonconformity score suggests a bad prediction of $f^c(\Bar{X}_k)$.

To obtain $\Bar{S}$, the nonconformity scores $S_{1}, \dots, S_{\bar{K}}$ are assumed as independently and identically distributed (i.i.d.) real-valued random variables, which allows permutating element of the set $\mathcal{S} = \{ S_{1}, \dots, S_{\bar{K}} \}$ (assumption of exchangeability) \cite{papadopoulos2002inductive}. Then, the set $\mathcal{S}$ is sorted in a non-decreasing manner expressed as $\mathcal{S} = \{S^{(1)},\dots, S^{(r)}, \dots, S^{(\overline{K})} \}$ where $S^{(r)}$ is the $(1-\alpha^c)$ sample quantile of $\mathcal{S}$. $r$ in $S^{(r)}$ is defined as $r:= \lceil (\overline{K}+1)(1-\alpha^c) \rceil$ where $\lceil \cdot \rceil$ is the ceiling function. Let $\Bar{S} = S^{(r)}$, then the following probability holds \cite{papadopoulos2002inductive},
\begin{equation}\label{eq: preli-cp}
	\mathrm{Pr}(f^c(X^{*})-S^{(r)} \leqslant Y^{*} \leqslant f^c(X^{*})+S^{(r)}) \geqslant 1-\alpha^c
\end{equation}

Thus, we can utilize $S^{(r)}$ to certify the uncertainty region over the predicted value of $Y^{*}$ given the prediction model $F$. It is noted that $\mathrm{Pr}(f^c(X^{*})-S^{(r)} \leqslant Y^{*} \leqslant f^c(X^{*})+S^{(r)}) \in [1-\alpha^c, 1-\alpha^c+\frac{1}{1+\overline{K}})$ \cite{papadopoulos2002inductive} and thus the probability in the Eq.~(\ref{eq: preli-cp}) holds.

\textbf{Adaptive Conformal Prediction.} CP relies on the assumption of exchangeability \cite{papadopoulos2002inductive} about the elements in the set $\mathcal{S}$. However, this assumption may be violated easily in dynamical time series prediction tasks \cite{gibbs2021adaptive}. To achieve reliable dynamical time series prediction in a long-time horizon, the notion of ACP\cite{gibbs2022conformal} was proposed to re-estimate the quantile number online. The estimating process is shown below:
\vspace{-0.3cm}
	\begin{equation}
		\label{eq: ACP update}
		\alpha_{k+1} = \alpha_{k} + \delta (\alpha^c-e_k) \; \mathrm{with}\; e_k=
		\begin{aligned}
			\begin{cases}
				0, \mathrm{if}\; S_{k} \leqslant S_k^{(r)}, \\
				1, \text{otherwise}.
			\end{cases}
		\end{aligned}
	\end{equation}

where $\delta$ is the user-specified learning rate. $S_k \in \mathbb{R}^{+}$ is the nonconformity score at time step $k$, and $S_k^{(r)}$ is the sample quantile. It is noted that if $S_{k} \leqslant S_k^{(r)}$, then $\alpha_{k+1}$ will increase to undercover the region induced by $S_k^{(r)}$.

\section{Method}
\subsection{System Dynamics}
Consider the discrete-time dynamical system defined below with the robot state $x_k \in \mathcal{X} \subset \mathbb{R}^n$ and the control input $u_k\in \mathcal{U} \subset \mathbb{R}^m$.
\begin{equation}\label{eq: dynamics}
	x_{k+1} = \hat{f}(x_k, u_k) + d(x_k, u_k) + \epsilon_k 
\end{equation}
where $\hat{f}: \mathcal{X} \times \mathcal{U} \mapsto \mathbb{R}^{n}$ is the known nominal discrete dynamics. $d: \mathcal{X} \times \mathcal{U} \mapsto \mathbb{R}^{n}$ is the unmodelled unknown dynamics which may come from the imprecise parameter in dynamics or the uncertainty from the environment, and is required to be learned through the data-driven model like the GPs \cite{koller2018learning} or the linear model \cite{kakade2020information}. $\epsilon_k\in \mathbb{R}^n$ is the process noise.
Eq.~\eqref{eq: dynamics} can be used to express a family of nonlinear systems such as quadrotor \cite{wang2018safe} and the pendulum \cite{koller2018learning}.

A common way to model the unknown dynamics $d$ is GPs as in \cite{koller2018learning}.  However, this method suffers from scalability issues due to their cubic computational complexity $O(N^3)$ for the number of data points $N$. Note that QFF \cite{mutny2018efficient} provides a balance between the computation efficiency and approximation precision of the GPs, facilitating efficient posterior sampling for online control. In this paper, $d(x_k, u_k) = W^{*} \phi(x_k, u_k)$ is adopted where $W^{*} \in \mathbb{R}^{n\times P}$ is the vector of weights (parameters) of the linear model. $\phi(x_k, u_k)$ ($\phi: \mathcal{X}\times \mathcal{U}$ $\mapsto \mathbb{R}^P$) is the QFF.

Note that in the practical data collection procedure, it is hard to explicitly remove the influence of the process noise. Consequently, our model is to learn the $d^{\epsilon}(x_k, u_k) = d(x_k, u_k) + \epsilon_k$. This may lead to the covariance shift and limit the performance of learned dynamics. To address this challenge, the statistic-driven method ACP-based Control Barrier Functions is proposed under the discrete-time setting. Readers can refer to section \ref{sec: pr cbf} for more details.

\subsection{Safe Learning Objective}
To learn the unknown dynamics defined in Eq.~\eqref{eq: dynamics} while executing the primary control task, one widely used method is to involve the learning procedure in the optimization framework like \cite{brunke2022safe,curi2020efficient}. Similarly, we define the optimization problem as below,

\begin{equation}\label{eq: objective}
	\min_{\pi \in \Pi} \min_{W} J^{\pi} (x_0; c, W)=\min_{\pi \in \Pi} \min_{W} \mathbb{E} [\sum_{k=0}^{K-1} c(x_k, u_k) | \pi, x_0, W]
\end{equation}

where $\pi$ is the control policy and $c\in \mathbb{R}^{+}$ is the immediate cost function. $K$ is the total time horizon under each episode of learning. The goal of the MBRL is to learn the unknown dynamics $d$ parameterized by $W^{*}$ to achieve the near-optimal cost of the objective function. However, solving Eq.~\eqref{eq: objective} is often intractable \cite{curi2020efficient,kakade2020information} because the unknown dynamics and the policy need to be optimized simultaneously. Section \ref{sec: algorithm analysis} will elaborate on how to approximate the optimization problem.

Solving the problem defined in Eq.~\eqref{eq: objective} does not necessitate safety during the learning process. Therefore, a safety constraint must be incorporated into safe learning tasks. Considering the uncertainty from $d$ and $\epsilon_k$ in Eq.~\eqref{eq: dynamics}, as discussed in \cite{wang2018safe, luo2022sample}, the safety may be guaranteed in a probabilistic manner. Given Eq.~\eqref{eq: dynamics}, the safe learning problem can be formulated as,

   \begin{equation}\label{eq: safe learning opt}
	\begin{aligned}
		\min_{\pi \in \Pi} \min_{W} J^{\pi}(x_0; c, W) 
		\quad s.t. \quad \mathrm{Pr}(c^s(x_k) \geqslant 0) \geqslant 1-\alpha
	\end{aligned}
    \end{equation} 

where $\pi$ is the policy to control the robot safely. $\alpha \in (0, 1)$ is the failure probability of the safety constraint. $c^s:\mathbb{R}^n \mapsto \mathbb{R}$ is a state-dependent safety function with $c^s(x_k)\geqslant 0$ represents the safety constraint in safe reinforcement learning.

To learn the unknown dynamics $d$, greedy exploration is not sample efficient. To that end, we will model this exploration process as a multi-bandit problem \cite{kakade2020information} to balance the exploitation and exploration process. Therefore, the goal is to reduce the gap between the optimal performance ($J^{\pi^{*}}$) under the true dynamics and the accumulated cost under the exploration policy when episodically learning the unknown dynamics, which is defined as the regret $R_T$ below.

\begin{equation}\label{eq: regret}
    R_T :=  \sum_{t=0}^{T-1} \sum_{k=0}^{K-1} c(x_k^t, u_k^t) - \sum_{t=0}^{T-1} J^{\pi^{*}}(x_0;c)
\end{equation}

where $x_k^t \in \mathbb{R}^n$ is the robot state at time step $k$ under the episode $t$ and  $T$ is the total number of training episodes. However, quantifying the probabilistic safety constraint in Eq.~\eqref{eq: safe learning opt} is difficult since 1) covariance shift of data collection may lead to the inaccurate prediction $d$, and 2) the previous method depends on the covariance of GPs to quantify uncertainty \cite{cheng2019end, wang2018safe} while some models, e.g. deep learning, can not directly output the uncertainty. Hence, it is significant to robustly quantify the probabilistic constraint without depending on a particular model form. 

\subsection{Discrete-Time Control Barrier Function with Probability Guarantee} \label{sec: pr cbf}
Control Barrier Functions have proved to be a promising way to render the safety set forward invariant\cite{ames2019control}. To accommodate it in the discrete-time dynamical systems, discrete-time control barrier functions (DT-CBF) under deterministic systems has been introduced in \cite{agrawal2017discrete}. The DT-CBF is summarized as below,

\textbf{Lemma. 1}
\label{lemma: DT-CBF}
	Discret-time control barrier function (DT-CBF) \cite{agrawal2017discrete}. Given a discrete-time deterministic dynamics $x_{k+1}=f(x_k, u_k), \forall k\in \mathbb{N}$ with the robot state $x_k \in \mathbb{R}^n$, control input $u_k \in \mathbb{R}^m$, and the true continuous dynamics $f: \mathbb{R}^n \times \mathbb{R}^m \mapsto \mathbb{R}^n$, the DT-CBF guarantees that the robot always stay in a 0-superlevel set $\mathcal{C} \subset \mathbb{R}^n$, $\mathcal{C} \triangleq \{x_k\in \mathbb{R}^n | h(x_k) \geqslant 0\}$ where $h: \mathbb{R}^n \mapsto \mathbb{R}$ is a continuous function, if $x_0 \in \mathcal{C}$ holds. Given the function $h$ for discrete-time deterministic dynamics and corresponding $\gamma \in (0, 1]$, the point-wise set of control that belongs to the set $\mathcal{H}(x_k) = \{u\in \mathbb{R}^{m} | h(f(x_k, u_k)) - h(x_k) \geqslant -\gamma h(x_k) \}$ enforces the safe set $\mathcal{C}$ forward invariant. 

However, the DT-CBF constraint may be uncertain due to imprecise dynamics, and hence we formulate the uncertainty-aware DT-CBF constraint in a probabilistic way as below,

\begin{equation}\label{eq: prob B}
    \mathrm{Pr}(h(\hat{f}(x_k, u_k) + d(x_k, u_k) + \epsilon_k) - h(x_k) \geqslant -\gamma h(x_k)) \geqslant 1-\alpha
\end{equation}

The uncertainty in Eq.(\ref{eq: prob B}) is from the prediction model describing $d$ and the process noise $\epsilon_k$. 

Previous method \cite{wang2018safe, cheng2019end} only considers the uncertainty from $d$ quantified by GPs. This paper considers the DT-CBF constraint as a prediction model and then constructs the nonconformity score online through ACP, which can quantify the uncertainty in Eq.(\ref{eq: prob B}). The benefit of ACP is that quantifying the uncertainty is independent of the prediction model choice. The following will present the details to quantify the uncertainty in Eq.(\ref{eq: prob B}).

Let $B(x_k, u_k)\triangleq h(\hat{f}(x_k, u_k) + d^{\epsilon}(x_k, u_k)) - (1-\gamma)h(x_k)$, where $B(x_k, u_k)$ is considered as a prediction model since $d^{\epsilon}(x_k, u_k)$  is the surrogate model (GPs) to predict unknown dynamics. 

Let $B^{*}(x_k, u_k) \triangleq h(\hat{f}(x_k, u_k) + d(x_k, u_k) + \epsilon_k) - (1-\gamma)h(x_k)$ represent the true value of prediction model $B(x_k, u_k)$. From Eq.(\ref{eq: preli-cp}) in section \ref{sec: ACP}, we know that:
\begin{align}\label{Eq: prob B ACP expand}
    \mathrm{Pr}( -S^{(r)} + B(x_k, u_k) \leqslant B^{*}(x_k, u_k) \leqslant S^{(r)} &+ B(x_k, u_k) ) \notag \\
    &\geqslant 1-\alpha
\end{align}

where $S^{(r)}$ is the quantile number which is obtained by the nonconformality score set $\mathcal{S}$ consisting of the nonconformality scores $S_k=|B^{*}(x_k, u_k) - B(x_k,u_k)|$ at each time step.

\textbf{Theorem. 1} Suppose there exists a failure probability $\alpha \in (0, 1)$. Given the system in Eq.(\ref{eq: dynamics}) and considering the learning time horizon $K$ with the quantile number $S^{(r)}$ where $r:= \lceil (K+1)(1-\alpha^c) \rceil$, quantifying Eq.(\ref{eq: prob B}) with $1-\alpha$ probability guarantee leads to the constraint,

    \begin{equation} \label{eq: prob B to B}
        Eq.(\ref{eq: prob B}) \Rightarrow -S^{(r)} + B(x_k, u_k) \geqslant 0
    \end{equation}

\begin{proof}
    Given the definition of $B(x_k, u_k)$, our goal is to guarantee Eq.(\ref{eq: prob B}), i.e., $\mathrm{Pr}(B^{*}(x_k, u_k)\geqslant 0) \geqslant 1-\alpha$. It is noted that the true prediction value can not be obtained at the current state. We only can utilize the prediction model $B(x_k, u_k)$ to estimate the region that $B^{*}(x_k, u_k)$ belongs to, i.e., $\mathrm{Pr}( -S^{(r)} + B(x_k, u_k) \leqslant B^{*}(x_k, u_k) \leqslant S^{(r)} + B(x_k, u_k) ) \geqslant 1-\alpha$ is satisfied by the ACP. $B^{*}(x_k, u_k)\geqslant 0$ is satisfied with $1-\alpha$ probability only through the lower bound that $-S^{(r)} + B(x_k, u_k) \geqslant 0$. 
\end{proof}

\section{Algorithm and Analysis} \label{sec: algorithm analysis}
\subsection{Calibrated Model}
Given the linear model mapped by the QFF, we will use ridge regression to avoid overfitting. The ridge regression model for solving the unknown dynamics $d$ is formulated below,
{\footnotesize
    \begin{equation}
	\overline{W}^t=\argmin_{W} \sum_{i=1}^{t-1} \sum_{k=1}^{K} || W \phi(x_k^i, u_k^i) - d(x_k^i, u_k^i) ||_2^{2} + \lambda ||W||_F
\end{equation}
}

where $K$ represents the total time step of each episode and $t$ denotes the number of training episodes so far. $||W||_F$ is the Frobenius norm of the weight $W\in \mathbb{R}^{P}$ where $P$ is the number of the features. At each time step $k$ under episode $i$, we can observe the unknown dynamics by $d^{\epsilon}(x_k^i, u_k^i) = x_{k+1}^i - \hat{f}(x_k^i, u_k^i)$. With that, the analytical solution of the ridge regression can be obtained by,

\begin{equation}\label{eq: regression solution}
	\begin{aligned}
		\Sigma^t = \Sigma^0 + \sum_{i=1}^{t-1} \sum_{k=0}^{K} \phi(x_k^i, u_k^i) \phi^{\mathrm{T}} (x_k^i, u_k^i) \\
		\overline{W}^t= {(\Sigma^{t})}^{-1} \sum_{i=1}^{t-1} \sum_{k=1}^{K} \phi^{\mathrm{T}}(x_k^i, u_k^i) d^{\epsilon}(x_k^i, u_k^i)
	\end{aligned}	
\end{equation}
where $\Sigma^t \in \mathbb{R}^{P \times P}$ is the covariance matrix. $\Sigma^0 \in \mathbb{R}^{P \times P}$ is the initial covariance matrix, which is often chosen as $\lambda I$ followed by \cite{kakade2020information}. However, we can not utilize $\Sigma^0= \lambda I$ in the initial episode because we have no prior about the unknown dynamics. Using this estimated prior may break the bounded unknown dynamics assumption and hence we need a safe initial dataset pre-collected similar to \cite{luo2022sample}. Through the initial safe dataset, we can use the regression model in Eq.~\eqref{eq: regression solution} to obtain a reasonable initial guess about the $\Sigma^0$.

To balance the exploration and exploitation, the lower confidence-based continuous control ($\mathrm{LC}^3$) algorithm \cite{kakade2020information} will be adopted in this paper. Starting from each episode $i$ and given all the previous trajectories $\tau_i =\{(x_0^{i}, u_0^{i}), \dots,$ $(x_{K-1}^i, u_{K-1}^i), x_{K}^i \}$ from episode $i=0$ to $t-1$, Eq.~\eqref{eq: regression solution} is applied to obtain $\overline{W}^t$. $\Sigma^t$ describes the uncertainty of the weight $\overline{W}^t$, which is a ball geometrically, i.e. $\mathrm{BALL}^t$.

Given the $\mathrm{BALL}^t$ at the beginning of each episode $t$, Tomphson sampling formulated by $\widetilde{W}^t \sim \mathcal{N}(\overline{W}^t, (\Sigma^t)^{-1})$ is applied to sample the unknown dynamics $d$ during the exploration. The probability to sample $\widetilde{W}^t$ inside $\mathrm{BALL}^t$ involved with the initial safe dataset is $1-2\delta^s$ where $\delta^s \in [0, 1]$ is the failure probability to be inside the ball without considering the safe initialization. Interested readers can find details in \cite{luo2022sample}. 

Different from \cite{kakade2020information,luo2022sample}, this paper extends the finite feature space to the infinite feature space (GPs) which is intractable in the posterior sampling for online control. While QFF has a precise approximation and low-cost computation relative to the GPs. Besides, this paper considers a new simple safety quantification method that is independent of the model choice. 

\subsection{Optimism-based Safe Learning for Control}

\begin{algorithm}[t]

	\caption{MPPI-CBF-ACP}\label{alg: safe learning}
	\KwIn{ACP failure probability $\alpha$ and learning rate $\delta$, MPPI prediction horizon $H$, $\gamma$ in CBFs, total time steps $K$, total training episodes $T$, initial safe state $x_0$ and goal state $x_g$}
	\KwOut{Safe control policy at each episode}
	\BlankLine
	\While{$t \leq T$}{
		Initialization $x_0^t \gets x_0 $ \;
		Sample $\widetilde{W}^t \sim \mathcal{N}(\overline{W}^t, (\Sigma^t)^{-1})$ ; // Thompson Sampling for Exploration \\

		\While{$k < K$}{
			$\pi_k^t \gets \argmin_{\pi} J^{\pi} (x_k^t;c, \widetilde{W}^t)$ ; // MPPI for $u^{\mathrm{reference}}$ \\

			$u_k^t \gets Eq.~\eqref{eq: QCQP}$ \;
			$d^{\epsilon}(x_k, u_k^t) \gets$ $x_{k+1}- f(x_k, u_k^t)$ \;
		}
		$\overline{W}^{t+1}, \Sigma^{t+1} \gets Eq.~\eqref{eq: regression solution}$ \;
	}
\end{algorithm}

The algorithm of the optimism-based safe learning for control is presented in algorithm \ref{alg: safe learning}. Since solving the safe learning objective function is usually NP-hard, i.e. obtaining the optimal policy and optimal dynamics simultaneously, we assume the access to history trajectory to acquire the $\overline{W}^t$ and utilize the Model Predictive Path Integral (MPPI) \cite{williams2017information} to obtain the optimal policy $\pi^{*}$ as in \cite{kakade2020information}, which is in conjunction with the Thompson Sampling. 

Then, at training episode $t$ and time step $k$, we use $\bar{u}^t_{k|k}$ as a nominal control policy and map $u^{\mathrm{ref}}:=\bar{u}^t_{k|k}$ to the safety constraint proposed in this paper, which is formulated as,
    \begin{align}\label{eq: QCQP}
    &u^t_k = \argmin_u || u - u^{\mathrm{ref}}  ||^2 \\ 
    \quad s.t.  \quad &Eq.(\ref{eq: prob B to B}), \quad u_k\in[u_{\mathrm{min}}, u_{\mathrm{max}}] \notag
    \end{align}

where $u_{\mathrm{max}}, u_{\mathrm{min}}$ are the minimum and maximum limit of the control input, respectively. Given the algorithm \ref{alg: safe learning}, the regret bound of our algorithm is shown in the following Proposition,

\textbf{Proposition. 1}
	Assume the robot can access the oracle which means the data from the environment and bounded second moments of each step cost function under the control policy inputted is also assumed. Then, the regret bound of our algorithm is,
	\begin{equation}
		\label{eq: our regret}
		\mathbb{E} [R_T] \leqslant \widetilde{\mathcal{O}}(\sqrt{P(P+n+K)K^3 T})
	\end{equation}
	where $K$ is the total time steps under each episode and total $T$ is the episode for the robot to learn the unknown dynamics. $P$ is the number of the QFF's feature. $n$ is the dimension of the robot state. 

\begin{proof}
	Suppose the finite dimension of $\phi$ for QFF and $\phi$ is uniformly bounded with $|| \phi(x, u) ||_2 \leqslant A$.  We can use the Theorem 3 in \cite{luo2022sample} summarized from the $\mathrm{LC}^3$ in \cite{kakade2020information}, which is presented as,
    {\footnotesize
    \begin{equation} \label{eq: bound proof}
		\mathbb{\bar{E}} [R_T] \leqslant \widetilde{\mathcal{O}}(\sqrt{P(P+n+K)K^3 T} \log (1+\frac{A^2 || W^* ||_2^2}{\sigma^2}))
	\end{equation}
    }
    
where $\widetilde{\mathcal{O}}(\cdotp)$ ignores the logarithmic factors in $K$ and $T$. 
 Omitting the $\log$ term, the regret bound is obtained in Eq.~\eqref{eq: our regret}.
\end{proof}

\section{Simulation}

The simulations on the robot with single integrator dynamics and the inverted pendulum are conducted to validate the safe learning framework proposed in this paper. To demonstrate the computation efficiency of the proposed framework which is independent of the prediction model to learn the unknown dynamics, we compare it against Gaussian Processes under Radial Basis Function (RBF) kernel and linear regression with RFF which is utilized in $\mathrm{LC}^3$ \cite{kakade2020information}. The unmodeled dynamics utilized in MPPI is then sampled from the posterior of the surrogate model through Thompson Sampling. In the end, the control reference generated from MPPI is filtered by the ACP-based CBF in Eq.(\ref{eq: prob B to B}) to guide the exploration safely in the environment. After every particular exploration episode, the accumulated testing reward is computed to evaluate the performance and sample efficiency across the surrogate model. For a fair comparison, all parameters relevant to the framework presented in this paper are kept the same except where explicitly noted. Additionally, the simulation time steps $K$ are fixed under every episode.

\subsection{Mobile Robot with Single Integrator Dynamics}
\label{Sec: Robot}
\begin{figure}[t]
	\centering

	\includegraphics[scale=1, width=0.43\linewidth]{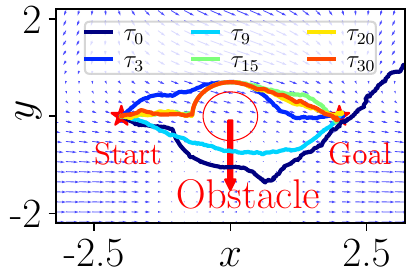}
    \quad
    \includegraphics[scale=1, width=0.45\linewidth]{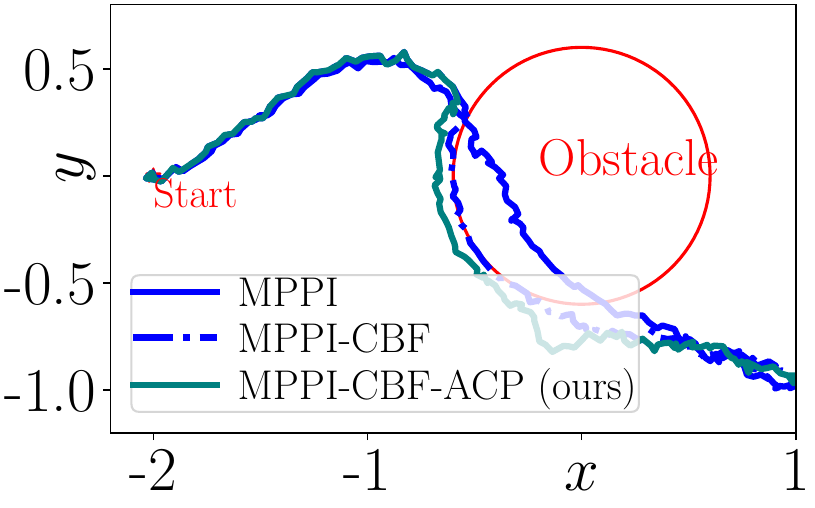}
	
	\caption{Simulation results on a mobile robot with single integrator dynamics. \textbf{Left:} The exploration path by our framework(QFF). $\tau{(\centerdot)}$ represents the episode index $(\centerdot)$ of the exploration path. \textbf{Right:} Ablation study to compare the safety performance from the exploration paths under three methods (dynamics is under the QFF model) respectively, i.e., 
 MPPI, MPPI-CBF, and MPPI-CBF-ACP (ours). }
	\label{Fig: integrator}
     \vspace{-0.5cm}
\end{figure}

The true dynamics of the single integrator is defined as, 
\begin{equation}\label{eq: dyn-single}
	x_{k+1} = x_k + u_k \varDelta t + d(x_k) + \epsilon_k
\end{equation}

where $x_k \in \mathbb{R}^2$ is the robot state. $u_k \in \mathbb{R}^2$ is the control input and  $\varDelta t = 0.01$. We follow \cite{berkenkamp2017safe,cheng2019end} to define the state-based $d$ as $d(x_k)=[0.01\sin{(x_k^{(2)})} \cos(x_k^{(1)}) + 0.006, 0.01e^{x_k^{(2)}}\cos{(3-0.5x_k^{(2)}})]^{\mathrm{T}}$ with $x_k=[x_k^{(1)}, x_k^{(2)}]^T$, which is shown in the Fig.~\ref{Fig: integrator}.
The motion noise is $\epsilon_k\sim \mathcal{N}(\mathbf{0}_2, \varDelta t^2 \mathbf{I}_{2\times2})$. The cost function to control the robot toward the goal is $c = (x - x_{\mathrm{goal}})^{\mathrm{T}} Q (x-x_{\mathrm{goal}}) + u^{\mathrm{T}} R u$ where $Q=\mathbf{I}_{2\times2}, R=\mathbf{I}_{2\times2}$. The safety set is $\mathcal{C}=\{\mathbf{x}_k | \mathbf{x}_k^T \mathbf{x}_k-0.6^2 \geqslant 0\}$. $\gamma=0.5$ and $\alpha=0.05$ are adopted. The total time step for each training episode is $K=300$. The feature $P=200$ for RFF and QFF is adopted. 

\textbf{Exploration validation and ablation study}. The simulation is conducted firstly to validate the effectiveness of the Thompson Sampling. We use QFF as a surrogate model to learn the unknown dynamics. The exploration path in Fig.\ref{Fig: integrator} qualitatively illustrates that the Thompson Sampling can guide the robot to explore the environment.

We continue to do the ablation study to evaluate the performance of our CBF-ACP by comparing it with MPPI, MPPI-CBF on a robot with single integrator dynamics. 
The simulation result in Fig.\ref{Fig: integrator} shows that the robots under algorithms MPPI and MPPI-CBF both collide with the obstacle while our method (CBF-ACP) can maintain safety.

\textbf{Safety validation.} To validate our method that is independent of the model choice to quantify the uncertainty, GPs, RFF, and QFF are applied in the simulation. At each time step, $h(x_k)$ is computed for safety evaluation.
After one episode of learning, the minimum $h(x^t_k)$ is stored. The value in Table \ref{tab: safety} is the smallest $h(x^t_k)$ with all the training episodes. Table \ref{tab: safety} shows that the proposed framework with different surrogate models is collision-free when exploring the environment, which further validates that our framework is model-independent to quantify the uncertainty.

\begin{figure}[t]
	\centering
    \includegraphics[scale=1, width=0.49\linewidth]{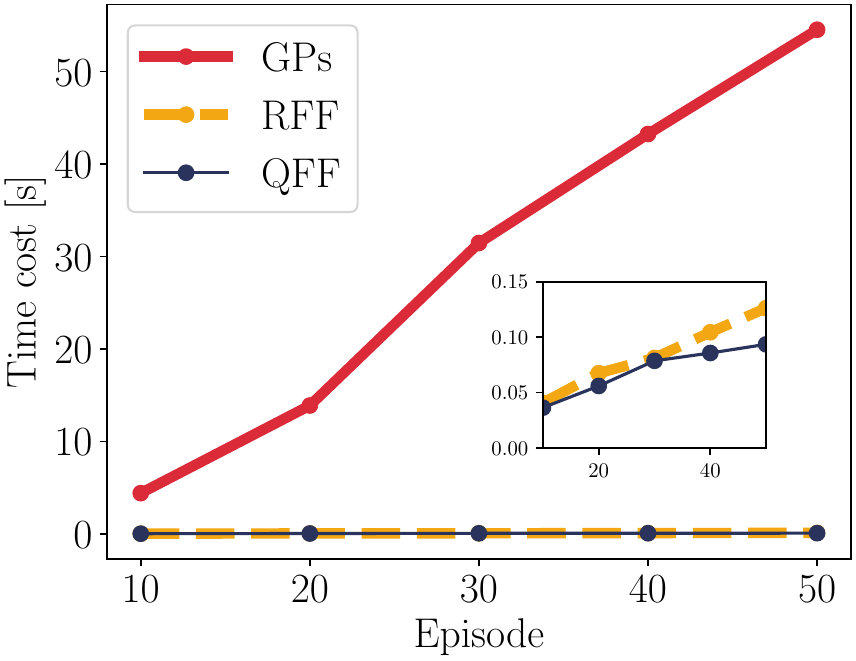}
	\includegraphics[scale=1, width=0.49\linewidth]{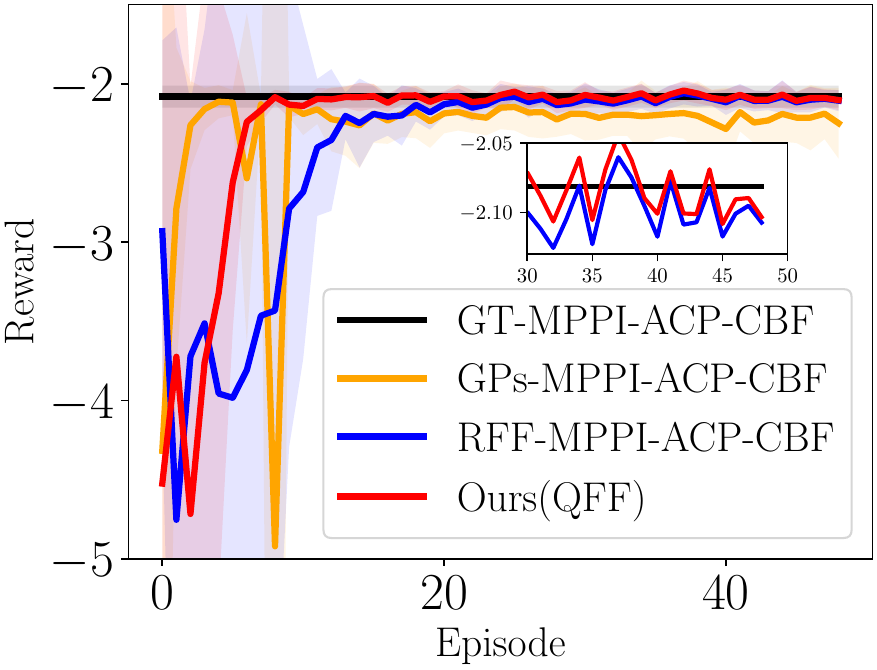}
	\caption{\textbf{Computation and sample efficiency evaluation.} \textbf{Left:} Training time for the learning process from examples in Fig.~\ref{Fig: integrator}. GP is not scalable as the number of training episodes increases. The time cost is validated on the i9-10900X CPU and RTX A5000 Graphics Card.
    \textbf{Right:} Reward for the pendulum under four models (GT-ground truth dynamics). The converge speed of our method is close to GPs and enjoys a better reward.}
	\label{Fig: performance}
\end{figure}

\textbf{Computation efficiency.} The training time cost for GPs, RFF, and QFF is shown in Fig.\ref{Fig: performance}, which illustrates that RFF and QFF can scale w.r.t. the input data size while GPs can not. Considering the expensive computation cost for GPs under episodic training, RFF and QFF are an alternative. It is noted that our framework guarantees safety with GPs.

\subsection{The Inverted Pendulum}
\begin{figure}[t]
	\centering
	\includegraphics[scale=1, width=0.4\linewidth]{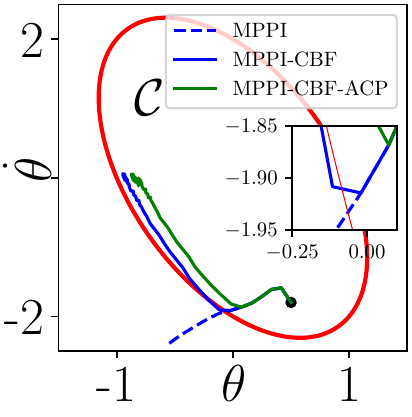}
	\quad
	\includegraphics[scale=1, width=0.4\linewidth]{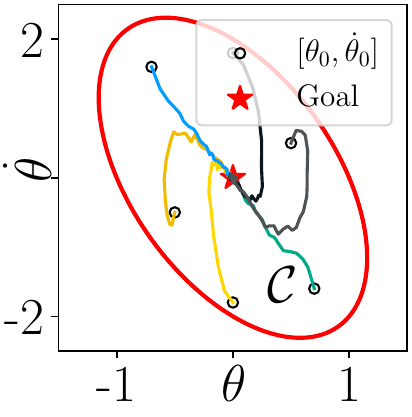}
	\caption{Pendulum simulation. \textbf{Left:} Ablation study to compare the safety performance from the exploration paths under three methods (dynamics is under the QFF model), i.e., MPPI, MPPI-CBF, and MPPI-CBF-ACP (ours). \textbf{Right:} Applying the learned dynamics for different initial states.}
	\label{Fig: pendulum}
\end{figure}

We adopt the inverted pendulum dynamics \cite{brockman2016openai} as below,
{\footnotesize\begin{equation}
	\begin{bmatrix}  
		\theta_{k+1} \\  
		\dot{\theta}_{k+1} 
	\end{bmatrix} =\begin{bmatrix}  
		\theta_{k} + \dot{\theta_{k}} \varDelta t + \frac{3g}{2l}\sin(\theta_{k}) (\varDelta t)^2  \\  
		\dot{\theta_{k}} + \frac{3g}{2l}\sin(\theta_{k}) \varDelta t 
	\end{bmatrix}+\begin{bmatrix}  
	\frac{3}{ml^2} (\varDelta t)^2 \\  
	\frac{3}{ml^2} \varDelta t 
	\end{bmatrix} u_k + \epsilon_k
\end{equation}}
where $\varDelta t =0.05$ and $m=l=1$ are applied. The angular velocity  $\dot{\theta}_k \in [-16, 16]$, the torque  $u_k \in [-5, 5]$, and $\epsilon_k \sim \mathcal{N}(\mathbf{0}_2, 0.005^2 \mathbf{I}_{2\times 2}))$ are employed. Similar to \cite{cheng2019end}, the safety set $\mathcal{C}$ is defined as,
\begin{equation}
    \mathcal{C} = \{ [\theta, \dot{\theta}]^{\mathrm{T}} \in \mathbb{R}^2 |
    1-\frac{\theta^2}{a^2} - \frac{\dot{\theta}^2}{b^2} - \frac{\theta \dot{\theta}}{ab} \geqslant 0
    \}
\end{equation}

where $a=1$ and $b=2$ are adopted. The safety set $\mathcal{C}$ is presented in Fig.\ref{Fig: pendulum}. For validating our algorithm, we use $m=1.2$ and $l=1.2$ to represent the imprecise estimation of the true pendulum dynamics, i.e. MPPI and CBF use the imprecise parameter. The cost function, $c=\theta^2 + 0.1\dot{\theta}^2 + 0.001 u^2$, is employed in the MPPI framework to keep the pendulum upright, i.e. $\theta=0$. The horizon in MPPI is 15 and the total time step for each training episode is 150.

\textbf{Safety validation.}
Similar to the simulation on the robot with single integrator dynamics, the framework under different surrogate models (GPs, RFF, and QFF) are conducted. Table \ref{tab: safety} shows that our framework can guarantee safety without depending on the model choice. To tell the effectiveness of the proposed framework (CBF-ACP) from the CBF, the simulation under QFF-CBF-ACP and QFF-CBF are conducted. Result in Fig.\ref{Fig: pendulum} validates that our framework can promise safety while QFF-CBF and only QFF can not guarantee safety.

\textbf{Sample efficiency.} To quantitatively evaluate the exploration efficiency, the reward w.r.t each training episode is presented in Fig.\ref{Fig: performance}. GPs first reach the ground-truth reward and then QFF is the second while RFF is the last. All the algorithms converge. The reward of QFF(ours) is a little higher than RFF because QFF provides a higher approximation precision of GPs. However, the reward of GPs is worse than QFF and RFF.

\textbf{Initial points.} We also utilize the learned $d$ to control the pendulum upright under different starts. Fig.\ref{Fig: pendulum} illustrates that the learned $d$ can reach the goal successfully using the learned unknown dynamics.

\begin{table}
\centering
\caption{Safety Evaluation: Minimal $h$ of All Training Episodes}
    \vspace{-0.2cm} 

    \begin{tabular}{c|c|c|c} 
    \hline
               & GPs & RFF & QFF  \\ 
    \hline
    $h^{\mathrm{integrator}}(x)$ &  0.021$\pm$0.006  &  0.008 $\pm$ 0.002   &   0.009$\pm$0.003   \\ 
    \hline
    $h^{\mathrm{pendulum}}(x)$  & 0.34$\pm$ 0.1    & 0.017$\pm$ 0.007  & 0.009$\pm$0.005    \\
    \hline
    \end{tabular}\label{tab: safety}
    \vspace{0.1cm} 
     
    {\footnotesize
    \parbox{\linewidth}{
        $h(x)>0$ is satisfied under our framework (CBF-ACP) to quantify the uncertainty. We conduct 10 learning processes and each learning process utilizes 50 episodes. Minimal $h(x)$ is obtained through all the training episodes under each model.}
    }
    \vspace{-0.5cm} 
\end{table}

\section{Conclusion}
This paper presents a computationally and sample efficient episodic safe learning method with certifiable high-probability safety guarantees.
Safety-critical constraints through adaptive conformal prediction are formulated to ensure safety throughout the learning process for the considered stochastic dynamical system. This constraint is embedded in an optimism-based exploration framework that is model-independent and effectively addresses the issue of covariance shift during data collection. Simulation results are provided to demonstrate the effectiveness and efficiency of the proposed method in safe learning tasks. Future work includes extending the proposed method to real-world experiments.

\bibliographystyle{IEEEtran}
\bibliography{IEEEabrv,ref}

\end{document}